\documentclass[runningheads]{llncs}
\usepackage[T1]{fontenc}
\usepackage{graphicx}
\usepackage{booktabs}
\usepackage{multirow}
\usepackage{arydshln}
\usepackage{xspace}
\usepackage{amsmath}
\usepackage{amssymb}
\usepackage[colorlinks]{hyperref}
\usepackage{dirtytalk}
\usepackage{caption}
\usepackage{subcaption}
\usepackage{float}
\usepackage{color}

\newcommand{\ghrepo}{https://github.com/sfu-mial/LesionElevation}

\usepackage{color}

\newcommand{\camready}[1]{{#1}}

\DeclareMathOperator{\argmax}{arg\,max}

\begin{document}
\title{Lesion Elevation Prediction from Skin\\Images Improves Diagnosis}
\author{
Kumar Abhishek\orcidID{0000-0002-7341-9617} \and
Ghassan Hamarneh\orcidID{0000-0001-5040-7448}
}
%
\authorrunning{
Abhishek and Hamarneh
}
\institute{
School of Computing Science, Simon Fraser University, Canada
\email{\{kabhishe,hamarneh\}@sfu.ca}
}
\maketitle              
\begin{abstract}
While deep learning-based computer-aided diagnosis for skin lesion image analysis is approaching dermatologists' performance levels, there are several works showing that incorporating additional features such as shape priors, texture, color constancy, and illumination 
further improves the lesion diagnosis performance. In this work, we look at another clinically useful feature, skin lesion elevation, and investigate the feasibility of predicting and leveraging skin lesion elevation labels. Specifically, we use a deep learning model to predict image-level lesion elevation labels from 2D skin lesion images.
We test the elevation prediction accuracy on the derm7pt dataset, and use the elevation prediction model to estimate elevation labels for images from five other datasets: ISIC 2016, 2017, and 2018 Challenge datasets, MSK, and DermoFit.
We evaluate cross-domain generalization by using these estimated elevation labels as auxiliary inputs to diagnosis models, and show that these improve the classification performance, with AUROC improvements of up to 6.29\% and 2.69\% for dermoscopic and clinical images, respectively. The code is publicly available at \url{\ghrepo}.

\end{abstract}
\section{Introduction}

\begin{figure*}[ht!]
    \centering
        \includegraphics[width=\textwidth]{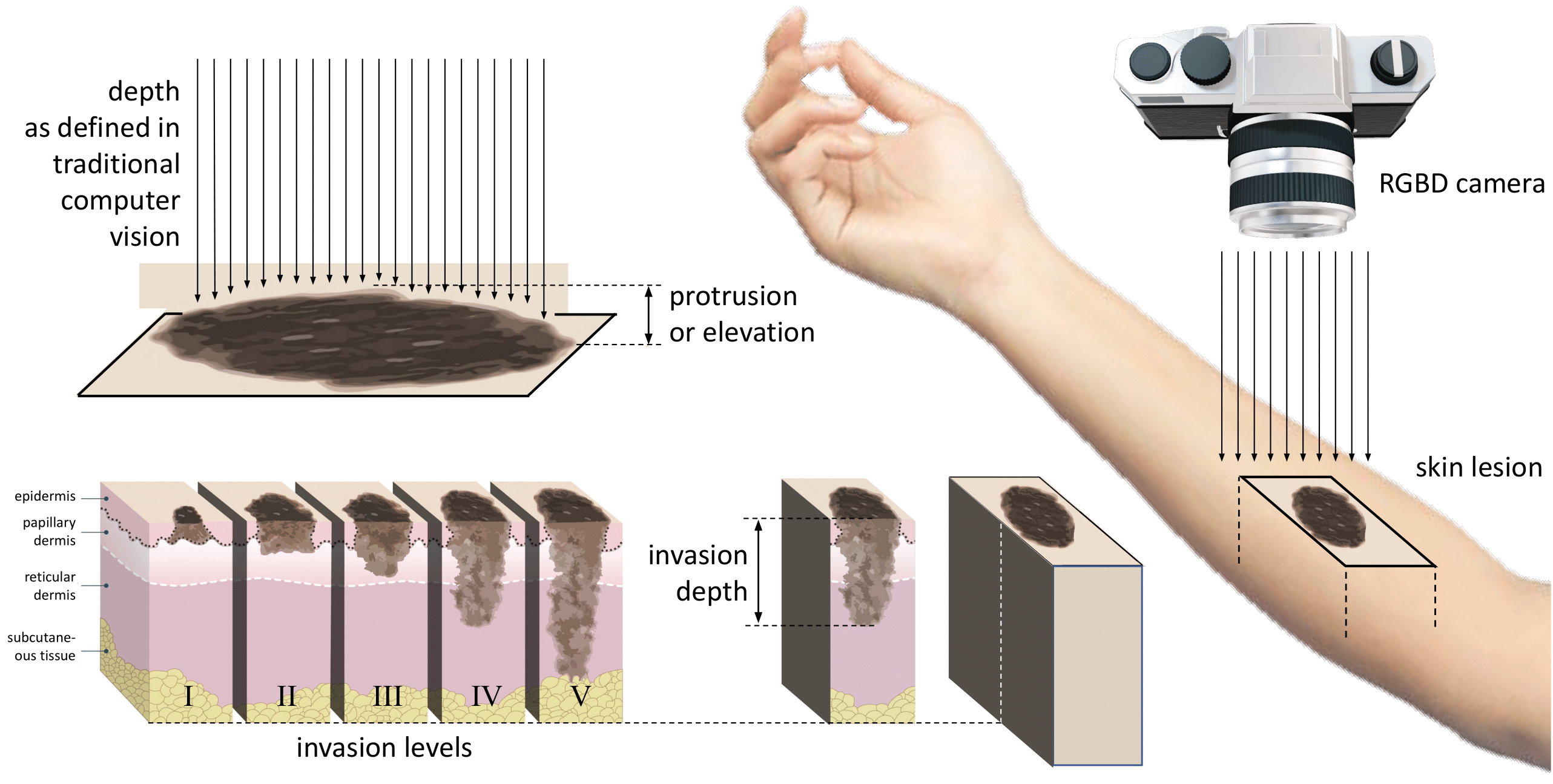}
        \caption{Visualizing the difference between skin lesion elevation versus depth. Invasion levels inset figure courtesy of Melanoma Institute Australia \cite{melanomadiagnosis}.
        }
        \label{fig:elevation_vs_depth}
\end{figure*}

Skin cancer is highly prevalent globally and the most commonly diagnosed cancer in the USA~\cite{FF2024} with over 5 million annual diagnoses~\cite{nagarajan2019keratinocyte}. 
Although it accounts for a small fraction of all skin cancers, melanoma is the deadliest form
with
an estimated
99,700 diagnoses and
8,290 deaths in 2024 in the USA alone,
and timely diagnosis is critical as early detection results in a 99\% estimated 5-year survival rate. Deep learning (DL)-based methods have proven to be successful in improving image-based clinical decision support systems with expert-level computer-aided dermatological diagnosis~\cite{Esteva2017,Brinker2019}. 
While DL-based methods have demonstrated remarkable performance, there is a considerable body of research showing that incorporating additional features, such as shape priors~\cite{mirikharaji2018star}, texture~\cite{zhang2019automatic}, color constancy~\cite{hua2019effect}, and illumination~\cite{abhishek2020illumination}, can further improve skin lesion image analysis. 
Another feature that has been shown to enhance lesion diagnosis and clinical management prediction, using both classical machine learning~\cite{li2009depth} and DL methods~\cite{kharazmi2018feature,Kawahara2019a,pacheco2020impact,abhishek2021predicting,mendes2022deep}, is lesion elevation.
However, learning-based methods have yet to incorporate lesion elevation prediction into computerized diagnosis.

The American Cancer Society's ABCDE criteria include elevation (E) as one of the components~\cite{strayer2003diagnosing}.
Moreover, in the clinical setting, dermatologists often palpate the skin to examine the lesion when making a diagnosis~\cite{cox2006palpation}. Case in point, a study showed that palpation alone, without any visual assessment, was sufficient to correctly diagnose 14 of 16 cases~\cite{cox2007literally}. 
With the rise of teledermatology, partly accelerated by external factors such as COVID-19~\cite{alabdulkareem2021palpation}, one of the major reasons for dermatologists' dissatisfaction with teledermatology is the inability to palpate lesions~\cite{eedy2001teledermatology,english2007has}. This is particularly pressing for \say{store-and-forward} teledermatology, where images are captured and submitted alongside patient history, which has been adopted for its efficiency and low cost, is imperfect
since \say{\emph{even good quality photos are two-dimensional; raised lesions $\ldots$ for example, may be difficult to distinguish from flat lesions of a similar colour}}~\cite{cox2006palpation}. Clinical and dermoscopic images of lesions do not capture elevation, and while it is recommended to capture tangential views of lesions in teledermatology, measuring elevation is not easy to do with limited camera views, making the examinations \say{less complete}~\cite{bashshur2015empirical,jahnke2022pediatric}.
Therefore, while teledermatology has the potential to improve triage, access to care for underserved communities, and patient convenience~\cite{morenz2019evaluation,coustasse2019use,santosa2023teledermatology,hwang2024review}, it would greatly benefit from being able to leverage lesion elevation information as a proxy for in-person palpation. 
Solutions to bridge this gap could be either in the form of patient side hardware~\cite{kim2016roughness}, which is expensive to develop, maintain, and deploy, or purely software-based approaches to estimate lesion elevation from single RGB images, which we focus on.

\begin{figure*}[ht!]
    \begin{subfigure}[t]{0.53\textwidth}
        \centering
        \includegraphics[width=\textwidth]{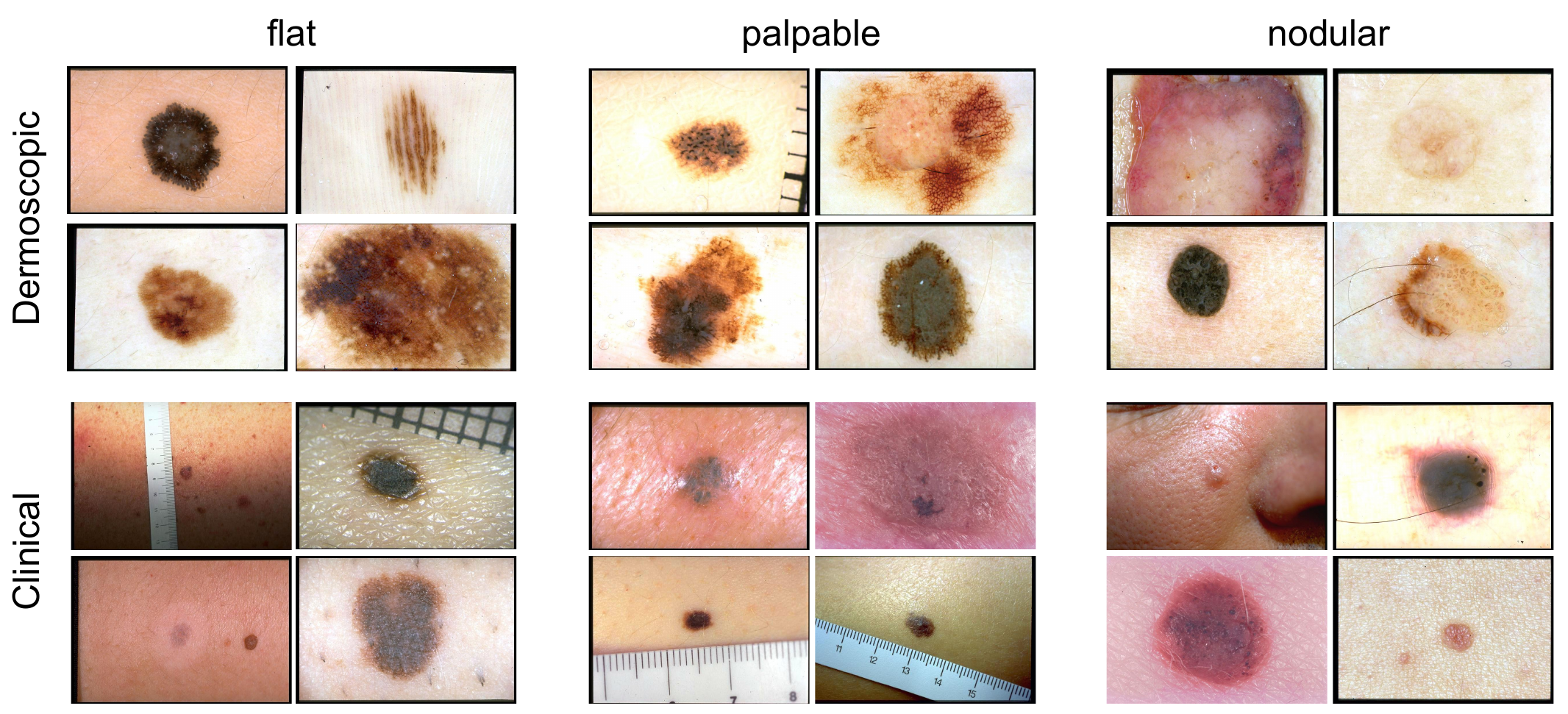}
        \caption{Sample dermoscopic and clinical images from derm7pt showing the three image-level lesion elevation labels.}
    \end{subfigure}
    \quad
    \begin{subfigure}[t]{0.42\textwidth}
        \centering
        \includegraphics[width=\textwidth]{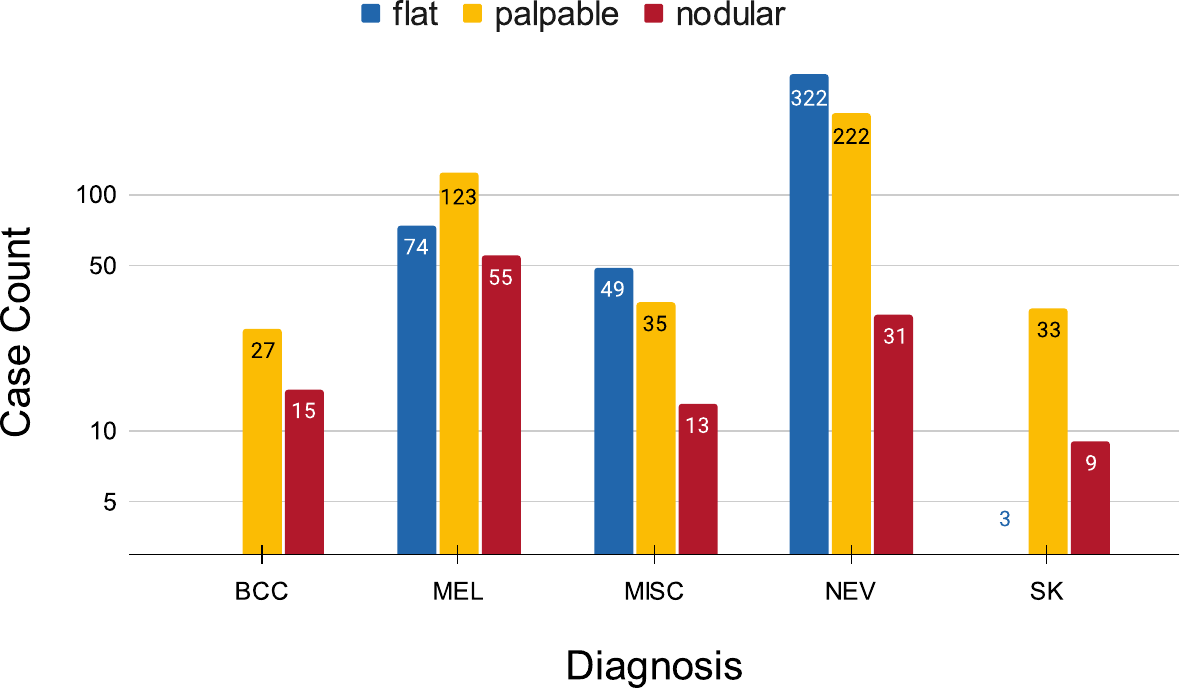}
        \caption{Distribution of elevation. Note the absence of a clear diagnosis-to-elevation mapping.}
    \end{subfigure}
    \caption{derm7pt dataset: (a) sample images categorized by elevation labels and (b) distribution of elevation labels across diagnoses.}
    \label{fig:dataset_overview}
\end{figure*}

Before proceeding, it is worth clarifying
the difference in terminology vis-à-vis lesion \say{elevation} and \say{depth} (Fig.~\ref{fig:elevation_vs_depth}), and how these terms differ in their usage in dermatology compared to traditional computer vision. 
Lesion elevation refers to the lesion's surface and how it protrudes above the outer skin surface (epidermis). On the other hand, lesion depth or thickness, unlike the definition of depth in computer vision, refers to the depth of invasion of melanoma underneath the skin surface and is used for melanoma staging, measured using scales such as Breslow's depth~\cite{breslow1970thickness} and Clark level~\cite{clark1969histogenesis}.

While lesion elevation has been used along with other clinical metadata (e.g., gender, lesion location, and age) for skin lesion image analysis tasks and has shown to improve performance~\cite{Kawahara2019a,abhishek2021predicting}, to the best of our knowledge, there is no work that explores either the utility of elevation alone as a metadata, or the feasibility of predicting elevation from 2D RGB skin lesion images. In this work, we pose three research questions: (i) can we predict, with sufficient accuracy, lesion elevation from a single lesion image?; (ii) does lesion elevation alone, without any other metadata, improve lesion diagnosis?; (iii) can we leverage an elevation prediction model to infer elevations on datasets without ground truth elevation, thus potentially improving the diagnosis accuracies thereon? Our results show that the answer is affirmative to all these questions.

\section{Method}

\noindent\textbf{The dataset:} Let $(\mathcal{X}, \mathcal{Y}, \mathcal{E})$ be the dataset of images, diagnosis labels, and elevation labels. Specifically, $\mathcal{X} = \{X_i\}_{i=1}^N$ is the set of skin images with corresponding diagnosis labels $\mathcal{Y} = \{Y_i\}_{i=1}^N$
and single image-level elevation labels $\mathcal{E} = \{E_i\}_{i=1}^N$,
where $X_i \in \mathbb{R}^{H \times W \times 3}$, $Y_i \in \{1, 2, \cdots, N_D\}$, 
and $E_i \in \{1, 2, \cdots, N_E\}$, and $N_D$ and $N_E$ denote the total number of diagnosis and image-level elevation class labels, respectively.

\noindent\textbf{A diagnosis prediction model:} A diagnosis prediction model $f_D$, parameterized by $\Theta_D$, is trained to generate disease predictions from images,
\begin{equation}
\label{eqn:diag_pred_vanilla}
    \widehat{Y}_i = f_D (X_i; \Theta_D).
\end{equation}

\noindent\textbf{Leveraging ground truth elevation labels for diagnosis prediction:} The $f_D$ model architecture can also be modified to take the elevation label as an auxiliary input for diagnosis prediction,
\begin{equation}
\label{eqn:diag_pred_with_gt_elev}
    \widehat{Y}_i = f_{DE} (X_i \oplus E_i; \Theta_{DE}),
\end{equation}

\noindent where $\oplus$ is a combination operator.

\noindent\textbf{Predicting elevation labels:} Additionally, since we have images with corresponding image-level elevation labels, we can also train a DL-model $g$ to predict elevation labels from an input image,
\begin{equation}
\label{eqn:infer_elevation}
    \widehat{E}_i = g (X_i; \Phi),
\end{equation}

\noindent where $\widehat{E}_i \in \mathbb{R}^{N_E}$ is an $N_E$-element probabilistic prediction of the elevation label. 
We denote elevation class label with the highest predicted probability as
$\widehat{E}_i^{\mathrm{max}} = \argmax_j \widehat{E}_{ij}$.
For example, if $N_E=3$ and $\widehat{E}_i = [0.1, 0.7, 0.2],$ then $\widehat{E}_{i}^{\mathrm{max}} = 2$.

\noindent\textbf{Leveraging predicted elevation labels for diagnosis models:} Finally, given a trained elevation prediction model $g$, we use this model to infer elevation labels (Eqn.~\ref{eqn:infer_elevation}) on datasets without ground truth elevation, and use these labels as auxiliary inputs to re-train the diagnosis prediction model,
\begin{equation}
\label{eqn:diag_pred_with_infr_elev}
    \widehat{Y}_i = f_{D\widehat{E}} (X_i \oplus \widehat{E}_i; \Theta_{D\widehat{E}}).
\end{equation}

More details about exact model architectures, losses, and metrics for evaluation are discussed in the next section.

\section{Results and Discussion}

\noindent\textbf{Datasets:} Since skin lesion elevation data is expensive and difficult to acquire, there are, to the best of our knowledge, only 2 publicly available datasets with lesion elevation labels: PAD-UFES-20 \cite{pacheco2020pad} and derm7pt \cite{Kawahara2019a}. PAD-UFES-20 contains \camready{2,298} smartphone camera images of skin lesions with binary labels indicating if a lesion is elevated or not, whereas derm7pt contains \camready{1,011 cases with} clinical and dermoscopic images with 3 elevation class labels: \say{flat}, \say{palpable}, and \say{nodular}, and because of the relatively more granular elevations and the presence of two imaging modalities in the latter, we use derm7pt for our experiments. We partition the dataset with 
elevation label-based
stratification into training, validation, and testing sets in the ratio of 70:15:15, accounting for the inherent class imbalance: \say{flat}: 448 cases, \say{palpable}: 440 cases, \say{nodular}: 123 cases. See Fig.~\ref{fig:dataset_overview} (a,b) for sample images from different elevation labels and diagnosis-wise distribution of elevation labels, respectively. We group the diagnosis labels in derm7pt into 5 classes as originally proposed by Kawahara et al.~\cite{Kawahara2019a}: BCC (basal cell carcinoma), MEL (melanoma), NEV (nevi), SK (seborrheic keratosis), and MISC (miscellaneous). 
\camready{Although some elevation labels appear more or less frequently with certain diagnoses, we}
note that there is no direct diagnosis-elevation mapping, and elevation labels are distributed across all diagnoses in our dataset (an exception is that BCC and SK have almost no \say{flat} elevations).

\begin{table}[ht!]
\centering
\caption{Results (accuracy and area under the ROC curve (for skin lesion elevation prediction from clinical and dermoscopic images of derm7pt. Reported values are the mean $\pm$ std. dev. averaged over 3 runs. Numbers in $[\cdot]$ present the 95\% CI values. Bold values denote the best values for the metrics.}
\resizebox{\textwidth}{!}{%
\setlength{\tabcolsep}{0.7em}
\def\arraystretch{1.35}
\begin{tabular}{@{}cc|cc|cc@{}}
\toprule
\multicolumn{2}{c}{Model}       & \multicolumn{2}{c}{Clinical Images}                                                                                                                               & \multicolumn{2}{c}{Dermoscopic Images}                                                                                                                            \\
    Architecture    & \# Params (M) & Accuracy $\uparrow$                                                                       & AUROC $\uparrow$                                                                           & Accuracy $\uparrow$                                                                        & AUROC $\uparrow$                                                                           \\ \midrule
MobileNetV2     &       2.228        & \begin{tabular}[c]{@{}c@{}}0.8234 $\pm$ 0.0515\\ \scriptsize{{[}0.7954, 0.8514{]}}\end{tabular} & \begin{tabular}[c]{@{}c@{}}0.7474 $\pm$ 0.0496\\ \scriptsize{{[}0.7251, 0.7697{]}}\end{tabular} & \begin{tabular}[c]{@{}c@{}}0.8039 $\pm$ 0.0552\\ \scriptsize{{[}0.7780, 0.8298{]}}\end{tabular} & \begin{tabular}[c]{@{}c@{}}0.7789 $\pm$ 0.0536\\ \scriptsize{{[}0.7549, 0.8029{]}}\end{tabular} \\ \hline
MobileNetV3L    &        4.206       & \begin{tabular}[c]{@{}c@{}}0.7969 $\pm$ 0.0561\\ \scriptsize{{[}0.7712, 0.8226{]}}\end{tabular} & \begin{tabular}[c]{@{}c@{}}0.7326 $\pm$ 0.0535\\ \scriptsize{{[}0.7111, 0.7541{]}}\end{tabular} & \begin{tabular}[c]{@{}c@{}}0.7908 $\pm$ 0.0576\\ \scriptsize{{[}0.7659, 0.8157{]}}\end{tabular} & \begin{tabular}[c]{@{}c@{}}0.7481 $\pm$ 0.0552\\ \scriptsize{{[}0.7260, 0.7702{]}}\end{tabular} \\ \hline
EfficientNet-B0 &       4.011        & \begin{tabular}[c]{@{}c@{}}0.8190 $\pm$ 0.0582\\ \scriptsize{{[}0.7914, 0.8466{]}}\end{tabular} & \begin{tabular}[c]{@{}c@{}}0.7444 $\pm$ 0.0570\\ \scriptsize{{[}0.7222, 0.7666{]}}\end{tabular} & \begin{tabular}[c]{@{}c@{}}0.8257 $\pm$ 0.0573\\ \scriptsize{{[}0.7978, 0.8536{]}}\end{tabular} & \begin{tabular}[c]{@{}c@{}}0.8088 $\pm$ 0.0563\\ \scriptsize{{[}0.7825, 0.8351{]}}\end{tabular} \\ \hline
EfficientNet-B1 &        6.517       & \begin{tabular}[c]{@{}c@{}}0.8013 $\pm$ 0.0604\\ \scriptsize{{[}0.7752, 0.8274{]}}\end{tabular} & \begin{tabular}[c]{@{}c@{}}0.7284 $\pm$ 0.0602\\ \scriptsize{{[}0.7071, 0.7497{]}}\end{tabular} & \begin{tabular}[c]{@{}c@{}}0.8344 $\pm$ 0.0604\\ \scriptsize{{[}0.8056, 0.8632{]}}\end{tabular} & \begin{tabular}[c]{@{}c@{}}0.8033 $\pm$ 0.0576\\ \scriptsize{{[}0.7774, 0.8292{]}}\end{tabular} \\ \hline
DenseNet-121    &       6.957        & \begin{tabular}[c]{@{}c@{}}0.8146 $\pm$ 0.0589\\ \scriptsize{{[}0.7874, 0.8418{]}}\end{tabular} & \begin{tabular}[c]{@{}c@{}}0.7405 $\pm$ 0.0568\\ \scriptsize{{[}0.7186, 0.7624{]}}\end{tabular} & \begin{tabular}[c]{@{}c@{}}0.8301 $\pm$ 0.0589\\ \scriptsize{{[}0.8018, 0.8584{]}}\end{tabular} & \begin{tabular}[c]{@{}c@{}}0.7931 $\pm$ 0.0544\\ \scriptsize{\scriptsize{{[}0.7680, 0.8182{]}}}\end{tabular}\\ \hline
VGG-16          &       14.724        & \begin{tabular}[c]{@{}c@{}}\textbf{0.8543 $\pm$ 0.0632}\\ \scriptsize{{[}0.8229, 0.8857{]}}\end{tabular} & \begin{tabular}[c]{@{}c@{}}\textbf{0.8220 $\pm$ 0.0610}\\ \scriptsize{{[}0.7941, 0.8499{]}}\end{tabular} & \begin{tabular}[c]{@{}c@{}}\textbf{0.8475 $\pm$ 0.0592}\\ \scriptsize{{[}0.8173, 0.8777{]}}\end{tabular} & \begin{tabular}[c]{@{}c@{}}\textbf{0.8152 $\pm$ 0.0582}\\ \scriptsize{{[}0.7883, 0.8421{]}}\end{tabular} \\    \hline
ResNet-18       &       11.178        & \begin{tabular}[c]{@{}c@{}}0.8190 $\pm$ 0.0582\\ \scriptsize{{[}0.7914, 0.8466{]}}\end{tabular} & \begin{tabular}[c]{@{}c@{}}0.7321 $\pm$ 0.0555\\ \scriptsize{{[}0.7106, 0.7536{]}}\end{tabular} & \begin{tabular}[c]{@{}c@{}}0.7996 $\pm$ 0.0536\\ \scriptsize{{[}0.7740, 0.8252{]}}\end{tabular} & \begin{tabular}[c]{@{}c@{}}0.7653 $\pm$ 0.0530\\ \scriptsize{{[}0.7422, 0.7884{]}}\end{tabular} \\ \hline
ResNet-50       &      23.514         & \begin{tabular}[c]{@{}c@{}}0.7660 $\pm$ 0.0607\\ \scriptsize{{[}0.7425, 0.7895{]}}\end{tabular} & \begin{tabular}[c]{@{}c@{}}0.6927 $\pm$ 0.0576\\ \scriptsize{{[}0.6732, 0.7122{]}}\end{tabular} & \begin{tabular}[c]{@{}c@{}}0.8083 $\pm$ 0.0576\\ \scriptsize{{[}0.7820, 0.8346{]}}\end{tabular} & \begin{tabular}[c]{@{}c@{}}0.7586 $\pm$ 0.0536\\ \scriptsize{{[}0.7359, 0.7813{]}}\end{tabular} \\
\bottomrule
\end{tabular}%
}
\label{tab:elevation_prediction}
\end{table}

In addition, we also use five other datasets for diagnosis prediction that do not contain ground truth elevation labels: (i) ISIC 2016~\cite{gutman2016skin}, (ii) ISIC 2017~\cite{codella2018skin}, (iii) ISIC 2018~\cite{codella2019skin}, (iv) MSK~\cite{ISICArchive}, and (v) DermoFit~\cite{ballerini2013color}, \camready{where (i) is a binary classification dataset and all others are multi-class classification datasets}. Note that (i)-(iv) are dermoscopic image datasets while (v) contains clinical images. For (i)-(iii), we use the standard dataset partitions, and for (iv), (v), we generate training, validation, and testing partitions in 70:10:20 ratio.

\smallskip
\noindent\textbf{Experiment 1: Can we predict skin lesion elevation labels from images alone?}
To test the feasibility of predicting skin lesion elevation labels directly from images, we train eight different DL model architectures on the derm7pt dataset. Specifically, we train elevation prediction models $g$ that, given a skin lesion image $X_i$, predict the elevation label $\widehat{E}_i$ (Eqn.~\ref{eqn:infer_elevation}). We choose the architectures from a variety of families, covering a large range of model sizes (see parameter counts in Table~\ref{tab:elevation_prediction}): ResNet-18 and ResNet-50, MobileNetV2 and MobileNetV3L, DenseNet-121, EfficientNetB0 and EfficientNetB1, and VGG-16. For all architectures except VGG-16, we modify the final layer to predict 3 classes ($N_E=3$ for derm7pt). However, since a large number of parameters in VGG-16 emanate from the fully-connected layers, we modify the architecture by replacing these fully-connected layers with a global average pooling layer~\cite{yan2019melanoma}. We use ImageNet-pretrained weights for initialization. All models are trained for 50 epochs with stochastic gradient descent and momentum of 0.9, weight decay of 1e-4, batch size of 32, and a learning rate of 1e-2 which was decayed by a factor of 0.1 every 10 epochs. All images are resized to $224 \times 224$ and we augment the images with horizontal and vertical flips and rotations in multiple of 90\textdegree.
To account for the inherent class imbalance, we use the cross-entropy loss with median frequency balancing to assign class weights, i.e. class-wise weights in the loss calculation are weighted by the ratio of the median of class frequencies in the entire training set to each class's frequency~\cite{eigen2015predicting,badrinarayanan2017segnet}. The model with the best area under the ROC curve (AUROC) on the validation set was chosen for evaluation. All experiments were repeated 3 times for robust results. 

\begin{figure}[t]
  \centering
    \includegraphics[width=\textwidth]{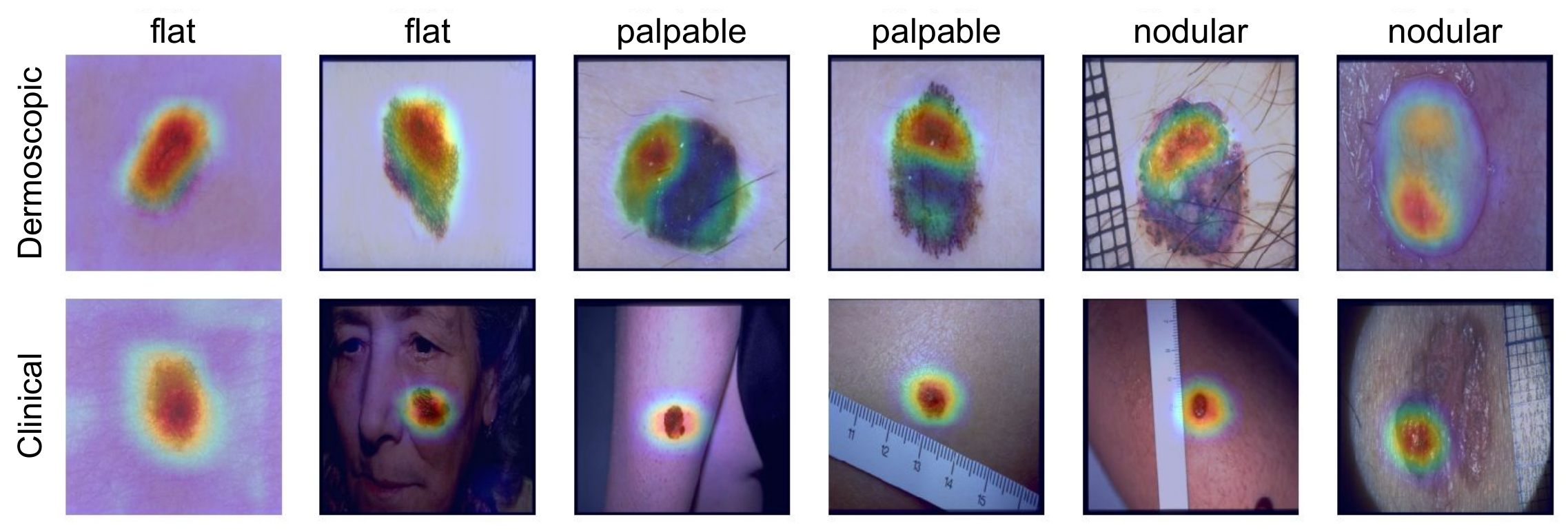}
    \caption{Visualizing class activation maps for skin lesion elevation label prediction for dermoscopic and clinical images, generated through GradCAM. Notice how the activation areas are focused around the lesion regions, indicating that the prediction model $g$ does not learn to rely on spurious features or ``shortcuts''.
    }
    \label{fig:gradcam}
\end{figure}

Table~\ref{tab:elevation_prediction} lists the quantitative results for elevation prediction on both clinical and dermoscopic images in the derm7pt dataset. We report mean and the std. dev. of the overall accuracy of classification as well as the AUROC across 3 repeated runs, as well as the 95\% confidence intervals (CIs). We observe that while all architectures are able to predict elevation labels reasonably accurately, the VGG-16 model performs the best across both imaging modalities. 
To ascertain that this performance is not due to the model learning spurious features or \say{shortcuts} in the images to make the predictions, we generate the class activation maps (CAMs) for the VGG-16 model using GradCAM~\cite{selvaraju2017grad}. Sample CAMs for both modalities and all elevation labels are shown in Fig.~\ref{fig:gradcam}. We observe that the CAMs are almost completely contained within and around the lesion regions, suggesting that the elevation predictions are indeed based on lesion features.
Since VGG-16 most accurately predicts elevation for both modalities, we use this model architecture for all subsequent experiments.

\smallskip
\noindent\textbf{Experiment 2: Do ground truth elevation labels help improve lesion diagnosis?} For this experiment, we train lesion diagnosis models $f_{DE}$ (Eqn.~\ref{eqn:diag_pred_with_gt_elev}) that leverage ground truth elevation labels as auxiliary inputs and compare their diagnosis performance to \say{vanilla} diagnosis models $f_D$ trained without any elevation labels (Eqn.~\ref{eqn:diag_pred_vanilla}). To combine the elevation labels as inputs along with the lesion image (i.e., the $\oplus$ operator in Eqn.~\ref{eqn:diag_pred_with_gt_elev}), we concatenate the one-hot encoded elevation labels for each image to the output of VGG-16's global-average pooling layer, which is then passed to the final classification layer, thus adding only a minimal number of parameters ($N_E \times N_D$, i.e., the number of elevation labels $\times$ the number of diagnosis classes). The training details (optimizer, loss, number of epochs, learning rate) for both $f_{DE}$ and $f_D$ remain the same.

\begin{table}[t]
\centering
\caption{
Leveraging inferred elevation labels (Eqn.~\ref{eqn:diag_pred_with_infr_elev}), either ``discrete'' ($f_{D\widehat{E}^{\mathrm{max}}}$) or ``probabilistic'' ($f_{D\widehat{E}}$) improves diagnosis performance over no elevation labels ($f_D$). Reported metrics are mean $\pm$ std. dev. over 3 repeated runs. We also report statistical significance tests (McNemar's mid-$p$ test) and effect sizes (Cohen's $d$). 
}
\label{tab:inferred_results}
\resizebox{\textwidth}{!}{%
\setlength{\tabcolsep}{0.7em}
\def\arraystretch{1.35}
\begin{tabular}{@{}cccccccccc@{}}
\toprule
\multirow{2}{*}{\textbf{Dataset}}   & \multirow{2}{*}{\textbf{Experiment}} & \multicolumn{6}{c}{\textbf{Metrics}}                                                                                                                                              & \multicolumn{2}{c}{\textbf{Statistical Tests}}                         \\ \cmidrule(l){3-10} 
                           &                             & \multicolumn{1}{c}{Bal. Acc. $\uparrow$} & \multicolumn{1}{c}{Accuracy $\uparrow$} & \multicolumn{1}{c}{Precision $\uparrow$} & \multicolumn{1}{c}{Recall $\uparrow$} & \multicolumn{1}{c}{F1-score $\uparrow$} & \multicolumn{1}{c}{AUROC $\uparrow$} & \multicolumn{1}{c}{$p$-value} & \multicolumn{1}{c}{Cohen's $d$} \\ \midrule
\multirow{3}{*}{DermoFit~\cite{ballerini2013color}}  & $f_D$                & $0.8145\pm0.0170$       & $0.9331\pm0.0051$        & $0.8121\pm0.0194$              & $0.8145\pm0.0170$           & $0.8103\pm0.0149$       & $0.8856\pm0.0092$          & -                            & -                              \\
                           & $f_{D\widehat{E}}$                 & $0.8586\pm0.0003$       & $0.9480\pm0.0003$        & $0.8449\pm0.0004$              & $0.8586\pm0.0003$           & $0.8511\pm0.0002$       & $0.9125\pm0.0001$          & 9.87e-03                     & 4.1348                         \\
                           & $f_{D\widehat{E}^{\mathrm{max}}}$             & $0.8541\pm0.0009$       & $0.9497\pm0.0010$        & $0.8466\pm0.0019$              & $0.8541\pm0.0009$           & $0.8500\pm0.0015$       & $0.9108\pm0.0007$          & 7.08e-03                     & 3.8626                         \\
                           \midrule
\multirow{3}{*}{MSK~\cite{ISICArchive}}       & $f_D$                & $0.6004\pm0.0010$       & $0.8446\pm0.0159$        & $0.6156\pm0.0302$              & $0.6004\pm0.0010$           & $0.5843\pm0.0091$       & $0.7374\pm0.0017$          & -                            & -                              \\
                           & $f_{D\widehat{E}}$                 & $0.6514\pm0.0019$       & $0.8833\pm0.0018$        & $0.7228\pm0.0037$              & $0.6514\pm0.0019$           & $0.6726\pm0.0013$       & $0.7747\pm0.0014$          & 4.04e-12                     & 23.9526                        \\
                           & $f_{D\widehat{E}^{\mathrm{max}}}$             & $0.6352\pm0.0047$       & $0.8878\pm0.0011$        & $0.7169\pm0.0210$              & $0.6352\pm0.0047$           & $0.6638\pm0.0021$       & $0.7632\pm0.0038$          & 4.10e-11                     & 8.7647                         \\
                           \midrule
\multirow{3}{*}{ISIC 2016~\cite{gutman2016skin}} & $f_D$                & $0.7008\pm0.0307$       & $0.8100\pm0.0474$        & $0.7208\pm0.0524$              & $0.7008\pm0.0307$           & $0.6998\pm0.0338$       & $0.7008\pm0.3070$          & -                            & -                              \\
                           & $f_{D\widehat{E}}$                 & $0.7344\pm0.0124$       & $0.8545\pm0.0131$        & $0.7615\pm0.0022$              & $0.7344\pm0.0124$           & $0.7467\pm0.0059$       & $0.7344\pm0.0124$          & 7.36e-02                     & 0.1547                         \\
                           & $f_{D\widehat{E}^{\mathrm{max}}}$             & $0.7574\pm0.0183$       & $0.8391\pm0.0165$        & $0.7513\pm0.0213$              & $0.7574\pm0.0183$           & $0.7515\pm0.0045$       & $0.7574\pm0.0183$          & 8.75e-02                     & 0.2603                         \\
                           \midrule
\multirow{3}{*}{ISIC 2017~\cite{codella2018skin}} & $f_D$                & $0.6926\pm0.0207$       & $0.8296\pm0.0072$        & $0.7303\pm0.0160$              & $0.6926\pm0.0207$           & $0.7060\pm0.0133$       & $0.6926\pm0.0207$          & -                            & -                              \\
                           & $f_{D\widehat{E}}$                 & $0.7417\pm0.0030$       & $0.8500\pm0.0060$        & $0.7634\pm0.0118$              & $0.7417\pm0.0030$           & $0.7513\pm0.0036$       & $0.7417\pm0.0030$          & 3.06e-02                     & 3.3198                         \\
                           & $f_{D\widehat{E}^{\mathrm{max}}}$             & $0.7555\pm0.0040$       & $0.8583\pm0.0044$        & $0.7776\pm0.0095$              & $0.7555\pm0.0040$           & $0.7644\pm0.0018$       & $0.7555\pm0.0040$          & 9.80e-03                     & 4.2192                         \\
                           \midrule
\multirow{3}{*}{ISIC 2018~\cite{codella2019skin}} & $f_D$                & $0.7949\pm0.0303$       & $0.9450\pm0.0055$        & $0.7601\pm0.0426$              & $0.7949\pm0.0303$           & $0.7690\pm0.0413$       & $0.8808\pm0.0190$          & -                            & -                              \\
                           & $f_{D\widehat{E}}$                 & $0.8481\pm0.0016$       & $0.9668\pm0.0021$        & $0.8314\pm0.0064$              & $0.8481\pm0.0016$           & $0.8390\pm0.0043$       & $0.9132\pm0.0014$          & 7.54e-03                     & 2.4051                         \\
                           & $f_{D\widehat{E}^{\mathrm{max}}}$             & $0.8524\pm0.0024$       & $0.9641\pm0.0032$        & $0.8250\pm0.0102$              & $0.8524\pm0.0024$           & $0.8376\pm0.0046$       & $0.9143\pm0.0012$          & 3.52e-03                     & 2.4885           \\
                           \bottomrule
\end{tabular}%
}
\end{table}

We observe that for clinical images, leveraging ground truth elevation labels for diagnosis prediction ($f_{DE}$) improves the performance [overall accuracy, AUROC]: [0.8569, 0.6820] compared to diagnosis without elevation ($f_D$): [0.8464, 0.6331]. A similar improvement is noted for dermoscopic images: the performance with elevation labels: [0.9216, 0.8703] is an improvement over that of a \say{vanilla} diagnosis model: [0.9137, 0.8431]. This improvement in AUROC of 4.89\% and 2.72\% for clinical and dermoscopic images, respectively, is consistent with findings from previous works~\cite{Kawahara2019a,abhishek2021predicting} that showed that using elevation labels along with other metadata is beneficial for lesion diagnosis prediction.

\smallskip
\noindent\textbf{Experiment 3: Can inferred elevation labels improve lesion diagnosis?} Having established that it is possible, with a reasonable accuracy, to predict elevation labels from lesion images, and that elevation labels improve lesion diagnosis, the natural next question is if we can infer lesion elevation on datasets that do not contain elevation labels, and if diagnosis prediction models trained with these inferred elevation labels also improve diagnosis accuracy. Therefore, given a trained elevation prediction model $g$, we infer elevation labels for all images in 5 skin lesion datasets that do not have elevation labels: ISIC 2016, ISIC 2017, ISIC 2018, MSK, and DermoFit. We note that there is a considerable domain shift between these skin lesion datasets~\cite{yoon2019generalizable}, and therefore we use modality specific elevation prediction models for inferring elevation labels, i.e., the elevation prediction model $g$ trained on derm7pt's dermoscopic images is used for the first 4 datasets, and $g$ trained on derm7pt's clinical images is used for DermoFit. Next, for each dataset, we train three prediction models: (i) diagnosis prediction without any elevation labels ($f_D$), (ii) diagnosis prediction with probabilistic \say{soft} inferred elevation labels ($f_{D\widehat{E}}$), and (iii) diagnosis prediction with \say{discrete} inferred elevation labels ($f_{D\widehat{E}^{\mathrm{max}}}$). Model training details remain the same as Experiment 1, except the models are trained for longer (20 epochs for ISIC 2018 and 50 epochs for the other datasets), since these datasets are larger than derm7pt. We report several classification metrics: balanced accuracy, overall accuracy, precision, recall, F1-score, and AUROC, and train each model thrice for robustness. In addition to these metrics, we also perform statistical analysis: McNemar's mid-$p$ test~\cite{mcnemar1947note,fagerland2013mcnemar} and effect size (Cohen's $d$~\cite{cohen2013statistical}) for comparing $\{f_{D\widehat{E}}, f_{D\widehat{E}^{\mathrm{max}}}\}$ AUROC predictions to those from $f_D$.

Quantitative results in Table~\ref{tab:inferred_results} show that leveraging estimated lesion elevation labels consistently improves diagnosis performance across all datasets: up to 6.29\% and 2.69\% improvements in AUROC for dermoscopic and clinical images, respectively. Moreover, for all datasets except ISIC 2016, this improvement is statistically significant at $p < 0.05$. Similarly, Cohen's $d$ estimates indicate \say{huge} effect sizes for these four datasets and \say{small} effect size for ISIC 2016~\cite{sawilowsky2009new}. While both \say{soft} and \say{discrete} estimates of the elevation label appear to improve diagnosis performance, interestingly, there does not appear to be a consistent pattern of one of them outperforming the other. This is especially surprising since the \say{soft} labels would convey the uncertainty associated with the elevation prediction, and intuitively, they would be more informative. Nevertheless, we shelve this observation for a future investigation.

\section{Conclusion}

In this work, we showed that it is possible to predict image-level lesion elevation labels directly from 2D RGB skin lesion images with sufficient accuracy, and that these estimated elevation labels do indeed help improve lesion diagnosis on other datasets, improving AUROC by up to 6.29\% and 2.69\% on dermoscopic and clinical images, respectively. The ability to predict lesion elevation from 2D images, in addition to improving computer-aided diagnosis, offers the potential to improve teledermatology consults by offering practitioners access to useful estimates of clinical information otherwise unavailable in virtual consultations.
Our experiments with off-the-shelf monocular depth prediction models~\cite{ranftl2022towards} from natural computer vision failed to generate any usable depth maps (see Fig. SM1 in the Supplementary Material), and we postulate that this may be because of the difference in scale of depth that these models are trained on (several orders of magnitude larger than skin lesion elevation) as well as the scene anisotropy of the images they are trained on (natural images generally have a depth anisotropy where the lower parts of the image are closer to the camera plane, which is typically not true for skin lesion images). Therefore, in future work, we would like to explore the feasibility, accuracy, and utility of reconstructing dense elevation maps from single RGB images, specific to skin lesions. Another future direction would be improving the elevation prediction accuracy, which may help reach the upper bound of performance improvement achieved when using ground truth elevation labels.
\camready{Finally, we would also like to explore using multiple datasets for training elevation labels' prediction models to alleviate any potential biases emanating from using a single dataset (derm7pt).}

\begin{credits}
\subsubsection{\ackname} 
The authors would like to thank Jeremy Kawahara, Manolis Savva, and Aditi Jain for helpful discussions on this work, and are grateful for the computational resources provided by NVIDIA Corporation and Digital Research Alliance of Canada (formerly Compute Canada). Partial funding for this project was provided by the Natural Sciences and Engineering Research Council of Canada (NSERC RGPIN/06752-2020).

\subsubsection{\discintname}
The authors have no competing interests to declare.
\end{credits}
%

%
%
\bibliographystyle{splncs04}
\bibliography{mybibliography_short}

\clearpage
\pagenumbering{arabic}
\setcounter{figure}{0}
\setcounter{footnote}{1}
\renewcommand{\thefigure}{SM\arabic{figure}}
\appendix

\section*{Supplementary Material}

All models were trained on an Ubuntu 20.04 workstation with AMD Ryzen 9 5950X, 32 GB of RAM, and NVIDIA RTX 3090 GPU, running Python 3.10.13 and PyTorch 2.1.2. The code is publicly available at \url{\ghrepo}.

\footnotetext{\url{https://github.com/isl-org/MiDaS/tree/79eda44b0c353f314070115d60215ae214227053}}

\begin{figure}[ht!]
  \centering
    \includegraphics[width=\textwidth]{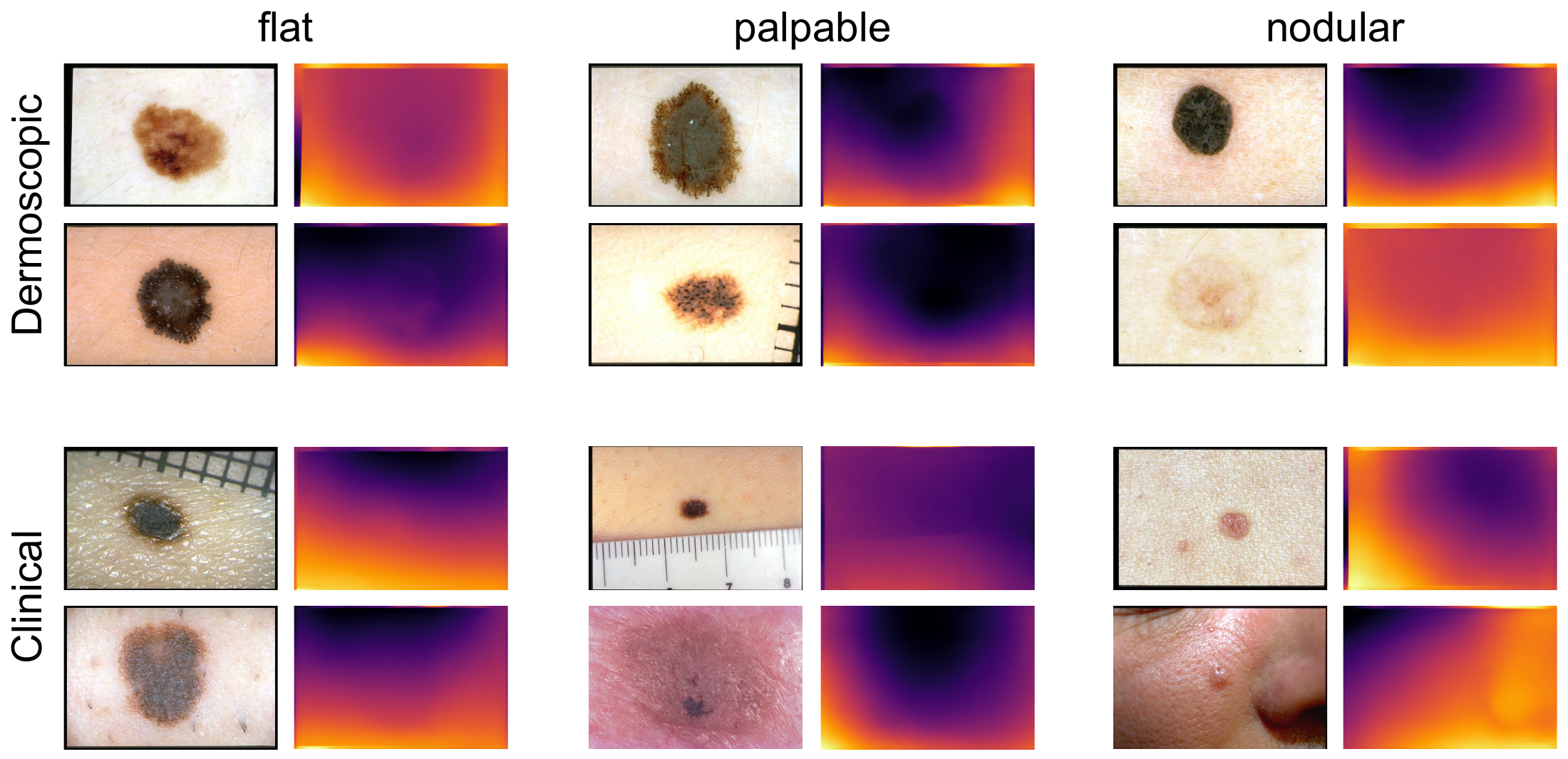}
    \caption{Dense depth maps for dermoscopic and clinical images from derm7pt generated using MiDaS~\cite{ranftl2022towards}, an off-the-shelf monocular depth estimation model\textsuperscript{1}. 
    Notice how the scene anisotropy of the images that this model has been trained on shows up in the generated depth maps - the lower regions of natural images are almost always closer to the camera plane, and this is reflected in these predicted depth maps. However, that is generally not the case with skin lesion images, which are typically acquired by having the camera plane parallel to the skin surface. This, along with the difference in scale between natural images' depths (typically in meters) and skin lesion elevations (typically in millimeters), considerably limits the utility of these depth map estimates.
    }
    \label{fig:midas_outputs}
\end{figure}
\end{document}